\documentclass[11pt,a4paper]{article}
\usepackage[hyperref]{emnlp-ijcnlp-2019}
\usepackage{times}
\usepackage{latexsym}

\usepackage{url}

\newcommand{\mn}[1]{{\textcolor{black}{#1}}}

\usepackage{enumitem}
\usepackage{color}
\usepackage{algorithm}
\usepackage[noend]{algpseudocode}
\usepackage{amsmath}
\usepackage{booktabs}
\usepackage{graphicx}
\usepackage{soul}

\makeatletter
\newcommand\footnoteref[1]{\protected@xdef\@thefnmark{\ref{#1}}\@footnotemark}
\makeatother

\aclfinalcopy 

\title{Machine Translation for Machines:\\ the  Sentiment Classification Use Case}

\author{Amirhossein Tebbifakhr$^{1,2}$, Luisa Bentivogli$^1$, Matteo Negri$^1$, Marco Turchi$^1$ \\
 $^1$ Fondazione Bruno Kessler, Via Sommarive 18, Povo, Trento - Italy \\
 $^2$ University of Trento, Italy\\
 {\tt \{atebbifakhr,bentivo,negri,turchi\}@fbk.eu}\\
} 

\date{}

\begin{document}
\maketitle
\begin{abstract}

We propose a neural machine translation (NMT) approach that, instead of pursuing adequacy and fluency (``human-oriented'' quality criteria), aims to  generate translations that are best suited as input to a natural language processing component designed for a specific downstream task (a ``machine-oriented''  criterion). Towards this objective, we present a reinforcement learning technique based on a  new candidate sampling strategy, which exploits  the results obtained on the downstream task as weak feedback. Experiments in sentiment classification of Twitter data in German and Italian show that feeding an English classifier with “machine-oriented” translations significantly improves its performance. 
Classification results outperform those obtained with translations produced by general-purpose \mn{NMT} models as well as by an approach based on reinforcement learning. 
\mn{Moreover, our results on both languages} approximate the classification accuracy computed on gold standard English tweets.
\end{abstract}

\section{Introduction}
\label{sec:intro}
Traditionally, machine translation (MT)  pursues a ``human-oriented'' objective: generating fluent and adequate output to be consumed by  speakers of the target language. But what if the intended use of MT is to feed a natural language processing (NLP) component instead of a human? This, for instance,  happens when MT is used as a pre-processing step to perform a downstream NLP task in a language for which dedicated tools are not available due to the scarcity of task-specific training data. The rapid growth of cloud-based software-as-a-service offerings  provides a typical example of this situation: a variety of affordable high-performance NLP tools can be easily accessed via  APIs but often they are available only for a few languages. Translating into one of these high-resource languages gives the possibility to address the downstream task by: \textit{i)} using  existing tools for that language to process the translated text, and 
\textit{ii)} projecting  their output back to  the original language.

However, using MT ``as is'' might not be optimal for different reasons. First, despite the qualitative leap brought by neural networks, MT is still not perfect \cite{koehn-knowles-2017-six}. Second, previous literature shows that MT can alter some of the properties of the source text \cite{mirkin-etal-2015-motivating,rabinovich-etal-2017-personalized,vanmassenhove-etal-2018-getting}. Finally, even in the case of a perfect MT able to preserve all the traits of the source sentence, models are still trained on parallel data, which are created by humans and thus reflect  quality criteria relevant for humans. 

In this work, we posit that these criteria might not be the optimal ones for a machine (i.e. a downstream NLP tool  fed with MT output). In this scenario, MT should pursue the objective of preserving and emphasizing those properties of the source text that are crucial for the downstream task at hand, even at the expense of human-quality standards. To this end, inspired by previous -- human-oriented -- MT approaches based on \textit{Reinforcement Learning} \cite{ranzato:2015,shen:2016} and \textit{Bandit Learning}  \cite{kreutzer:2017,nguyen:2017},  we explore a NMT optimization strategy that  exploits the weak feedback from the downstream task to influence system's behaviour towards the generation of optimal ``machine-oriented'' output. 

As a proof of concept, we test our approach on a sentiment classification task, in which Twitter data in German and Italian are to be classified according to their polarity by means of an English classifier.  In this setting, a shortcoming of previous translation-based approaches   \cite{Denecke08,balahur:2014}  is that, similar to other traits,  sentiment is often not preserved by MT \cite{salameh:2015,Mohammad:2016,lohar:2017}. 
\mn{Although it represents a  viable solution to  leverage sentiment analysis to a wide number of languages \cite{araujo:2016},  the translation-based approach should hence be supported by advanced technology able to preserve the sentiment traits of the input.}
\mn{Along this direction,} our experiments show that machine-oriented MT optimization makes the classifier's task easier and eventually results in significant classification improvements. Our results  outperform  those  obtained with translations produced by general-purpose  NMT  models  as  well  as  by  an NMT approach based on reinforcement 
learning \mn{\cite{ranzato:2015}.} Most noticeably, on both languages we are able to approximate the classification  accuracy computed on gold standard English tweets. 

\section{Background and Methodology}
\label{sec:method}

During training, NMT systems based on the encoder-decoder framework 
\mn{\cite{sutskever:2014, bahdanau:2014}}
are optimized with maximum likelihood estimation (MLE), which aims to maximize the log-likelihood of the training data. In doing so, they indirectly model the human-oriented quality criteria (adequacy and fluency) expressed in the training corpus. A different strand of research \cite{ranzato:2015,shen:2016,kreutzer:2018} focuses on optimizing   the  model parameters by maximizing an objective function that leverages either an evaluation metric like BLEU \cite{papineni2002bleu} or an external human feedback.  
\mn{These methods are based on \textit{Reinforcement Learning} (RL), in which}
the MT system parameters  $\theta$ define a \textit{policy} that chooses an \textit{action}, i.e. generating the next word in a translation candidate $\hat{\mathbf{y}}$, and gets a \textit{reward} $\Delta(\hat{\mathbf{y}})$ according to that action. Given $S$ training sentences $\{\mathbf{x}^{(s)}\}_{s=1}^S$, the RL 
training 
goal is to maximize the expected reward:

\begin{equation}
\begin{aligned}
    \mathcal{L}_{RL} & = \sum_{s=1}^{S} E_{\hat{\mathbf{y}} \sim     p_\theta(.|\mathbf{x}^{(s)})} \Delta(\hat{\mathbf{y}}) \\
           & = \sum_{s=1}^{S}\sum_{\hat{\mathbf{y}} \in \mathbf{Y}} p_\theta(\hat{\mathbf{y}}|\mathbf{x}^{(s)})  \Delta(\hat{\mathbf{y}})
\end{aligned}
\end{equation}
where $\mathbf{Y}$ is the set of all translation candidates.  Since the size of this set is exponentially large, it is impossible to  exhaustively
compute the expected 
reward, which is thus 
estimated by sampling one or few candidates from this set. In  the MT adaptation \cite{ranzato:2015} of 
REINFORCE \cite{Williams:1992}, the closest approach to the one we present, only one candidate is sampled. 

\begin{equation}
\begin{aligned}
\hat{\mathcal{L}}_{RL} = \sum_{s=1}^{S}p_\theta(\hat{\mathbf{y}}|\mathbf{x}^{(s)}) \Delta(\hat{\mathbf{y}}), \hat{\mathbf{y}} \sim p_\theta(.|\mathbf{x}^{(s)})
\end{aligned}
\end{equation}

 We now focus on how the key elements of  RL methods have been adapted to properly work in the proposed ``MT for machines'' setting. Our novel \textit{Machine-Oriented} approach is described in Algorithm \ref{algo:sampling}.

\begin{algorithm}[h]
\caption{\textit{Machine-Oriented} RL}\label{algo:sampling}
\begin{algorithmic}[1]
\State \textbf{Input:} $\mathbf{x}^{(s)}$ s-th source sentence in training data, $\mathbf{l}^{(s)}$ the ground-truth label, K number of sampled candidates 
\State \textbf{Output:} sampled candidate $\hat{\mathbf{y}}^{(s)}$
\State $\mathbf{C} = \emptyset$  \Comment{Candidates set}
\For{$k$ = 1,...,K}
 \State $\mathbf{c} \sim p_\theta(.|\mathbf{x}^{(s)})$ 
 \State $\mathbf{f} = P_{class}(\mathbf{l}^{(s)}|\mathbf{c})$ \Comment{Feedback from the classifier}
 \State $\mathbf{C} = \mathbf{C} \cup (\mathbf{c},\mathbf{f})$ 
\EndFor
\State $\hat{\mathbf{y}}^{(s)} = \max_{\mathbf{f}}(\mathbf{C})$ \Comment{Best candidate w.r.t. feedback}
\end{algorithmic}
\end{algorithm}

\paragraph{Reward Computation.}
In current RL approaches to MT, rewards are computed either on a development set containing reference translations \cite{ranzato:2015,bahdanau:2016} or by means of a weak feedback (e.g. a 1-to-5 score) when ground-truth translations are not accessible \cite{kreutzer:2017,nguyen:2017}. In both cases, the reward reflects a human-oriented notion of MT quality. Instead, in our machine-oriented scenario, the reward reflects the performance on the downstream task, independently from 
translation quality.

In the sentiment classification use case, given a source sentence ($\mathbf{x}^{(s)}$) and its sentiment ground-truth label ($\mathbf{l}^{(s)}$), the reward is defined as the probability assigned  by the classifier to the ground-truth class for each translated sentence \textbf{c} ($P_{class}(\mathbf{l}^{(s)}|\mathbf{c})$ in Algorithm \ref{algo:sampling}, line 6). 
By maximizing this reward, the MT system learns to produce a translation that has higher chances 
to be correctly labeled by the classifier. This comes at the risk of obtaining 
less fluent and adequate output. However, as will be shown in Section \ref{sec:results}, this type of reward induces highly polarized translations that are best suited for our downstream task.

\paragraph{Sampling Approach.} 
A possible sampling strategy is to exploit beam search \cite{sutskever:2014} to find, at each decoding step, the candidate with the highest probability.  Another solution is to use  multinomial sampling \cite{ranzato:2015} which, at each decoding step, samples tokens over the model's output distribution. In \cite{wu:2018}, the higher results achieved by multinomial sampling are ascribed to its capability to better explore the probability space by generating more diverse candidates.
This finding is particularly relevant in the proposed ``MT for machines'' scenario, in which the emphasis on 
final 
performance in the downstream task admits radical (application-oriented) changes in the behaviour of the MT model, even at the expense of human quality standards. 

To increase the possibility of such changes, we propose a new  sampling strategy.  Instead of  generating only one candidate \mn{\textit{token}} via multinomial sampling, $K$ 
\mn{candidate \textit{sentences}} 
are first randomly sampled (lines 4-5 in Algorithm 1). Then, 
the reward is 
collected for each of them (line 7) and the 
candidate 
with the highest reward is chosen (line 8). 
On one side, randomly exploring more candidates increases the probability to 
sample 
a ``useful'' one, possibly by diverting from the initial model behaviour.  On the other side, selecting the  candidate with the highest reward will push the system towards translations 
emphasizing
input traits that are relevant for the downstream task \mn{at hand}. In 
the sentiment classification use case, these are 
expected to be
sentiment-bearing terms that help the classifier to predict the correct class.

\section{Experiments}
\label{sec:experiments}

Our evaluation is done by  feeding an English sentiment classifier with the translations of German and Italian tweets generated by: 
\begin{itemize}
    \item A general-purpose NMT system (\textit{Generic});
    \item The same system conditioned with REINFORCE (\textit{Reinforce});
    \item The same system conditioned with our \textit{M}achine-\textit{O}riented method (\textit{MO-Reinforce}).
\end{itemize}

As other terms of comparison, we calculate the results of: 
\begin{itemize}
    \item The English classifier on the gold standard English tweets (\textit{English});
    \item German and Italian  classifiers on the original, untranslated tweets (\textit{Original}).
\end{itemize}

\paragraph{Task-specific data.}
We experiment with a dataset based on Semeval 2013 data \cite{semeval:2013},  which contains polarity-labeled parallel  German/Italian--English corpora \cite{balahur:2014}. 
For each language pair, the 
\mn{development}
and test sets respectively comprise 583  (197 negative and 386 positive) and 2,173 tweets (601 \mn{negative} and 1,572 \mn{positive}). To cope with the skewed data distribution, the negative tweets in the 
\mn{development} sets are over-sampled, leading to new balanced 
sets 
of 772 tweets.

\paragraph{NMT Systems.} 
Our \textit{Generic} models are based on Transformer \cite{vaswani:2017}, with parameters similar to those used in the original paper. Training data amount to 6.1M (De-En)
and 4.56M (It-En) parallel sentences from freely-available corpora.
The statistics of the parallel corpora are reported in Table \ref{tab:data}.
For each language pair, all data are merged and tokenized. Then,
\mn{byte pair encoding}
\cite{sennrich:2016} is applied to obtain 32K sub-word units.

\begin{table}[h]
    \centering
    \begin{tabular}{c|c|c}
                    \toprule
                    & De-En & It-En \\
                    \midrule
        Europarl    & 2M    & 2M    \\ 
        JRC	        & 0.7M  & 0.8M  \\
        Wikipedia   & 2.5M	& 1M    \\
        ECB	        & 0.1M  & 0.2M  \\
        TED	        & 0.1M	& 0.2M  \\
        KDE	        & 0.3M	& 0.3M  \\
        News11      & 0.2M	& 0.04M \\
        News	    & 0.2M	& 0.02M \\
                    \midrule
        Total       & 6.1M	& 4.56M \\
                    \bottomrule
        
    \end{tabular}
    \caption{Statistics of \mn{the} parallel corpora used for training the generic NMT systems}
    \label{tab:data}
\end{table}

To emulate both scarce and sufficient training data conditions, MT systems are trained using 5\%
and 100\% of the available parallel data.
\mn{In the most favorable condition (i.e. with 100\% of the data), 
the BLEU score of the two models is 30.48 for De-En and 28.68  for  It-En.}

To condition the generic models and obtain the \textit{Reinforce} and \textit{MO-Reinforce} systems, we use the polarity-labeled 
German/Italian tweets in
our 
\mn{development} sets, with reward as defined in \mn{Section}~\ref{sec:method}. The SGD optimizer is used, with learning rate \mn{set to} 0.01. In \textit{MO-Reinforce},  the number of sampled candidates is set to $K=5$.

\paragraph{Classifiers.}
To simulate an English cloud-based classifier-as-a-service, the pre-trained BERT (Base-uncased) model \cite{bert:2018} is fine-tuned with a balanced set of 1.6M positive and negative English tweets  \cite{go:2009}.
The German and Italian classifiers are also created by fine-tuning BERT
\mn{on the polarity-labeled tweets composing the source side of our development set (772 tweets).}\footnote{This strategy is similar to the one proposed in \cite{DBLP:journals/corr/abs-1809-04686}, where a pre-trained multilingual encoder is used to build a \mn{cross-lingual} 
classifier that is then trained on task-specific data. In our case, however, labeled data are available in a small amount.}
Before being passed to the classifiers, 
URLs and user mentions are removed from the tweets, which are then tokenized and lower-cased.

\section{Results and Discussion}
\label{sec:results}

\begin{figure*}
\centering
\begin{minipage}[b]{.46\textwidth}
\centering
\includegraphics[scale=0.5]{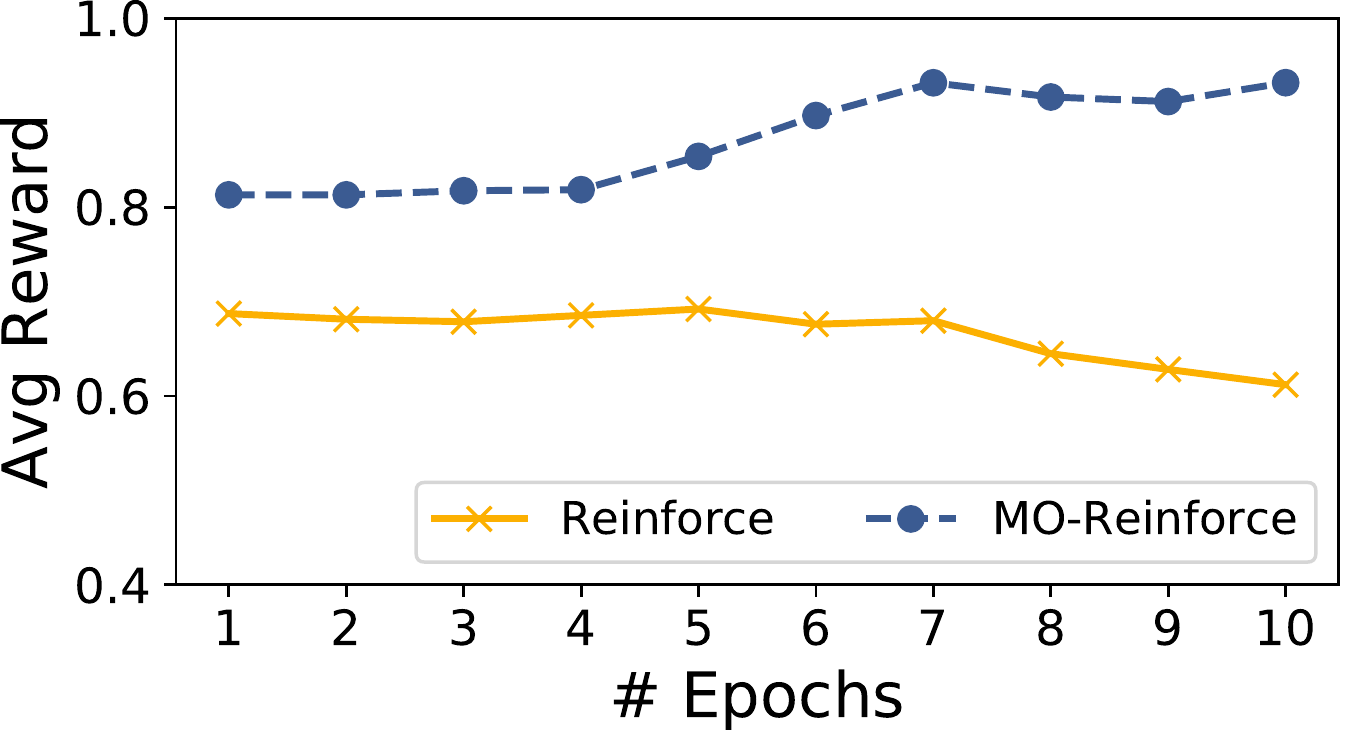}
\caption{Average rewards for \textit{Reinforce} and \textit{MO-Reinforce} candidates at each training epoch (De-En).}
\label{fig:avg_reward-DEEN}
\end{minipage}\qquad
\begin{minipage}[b]{.45\textwidth}
\centering
\includegraphics[scale=0.5]{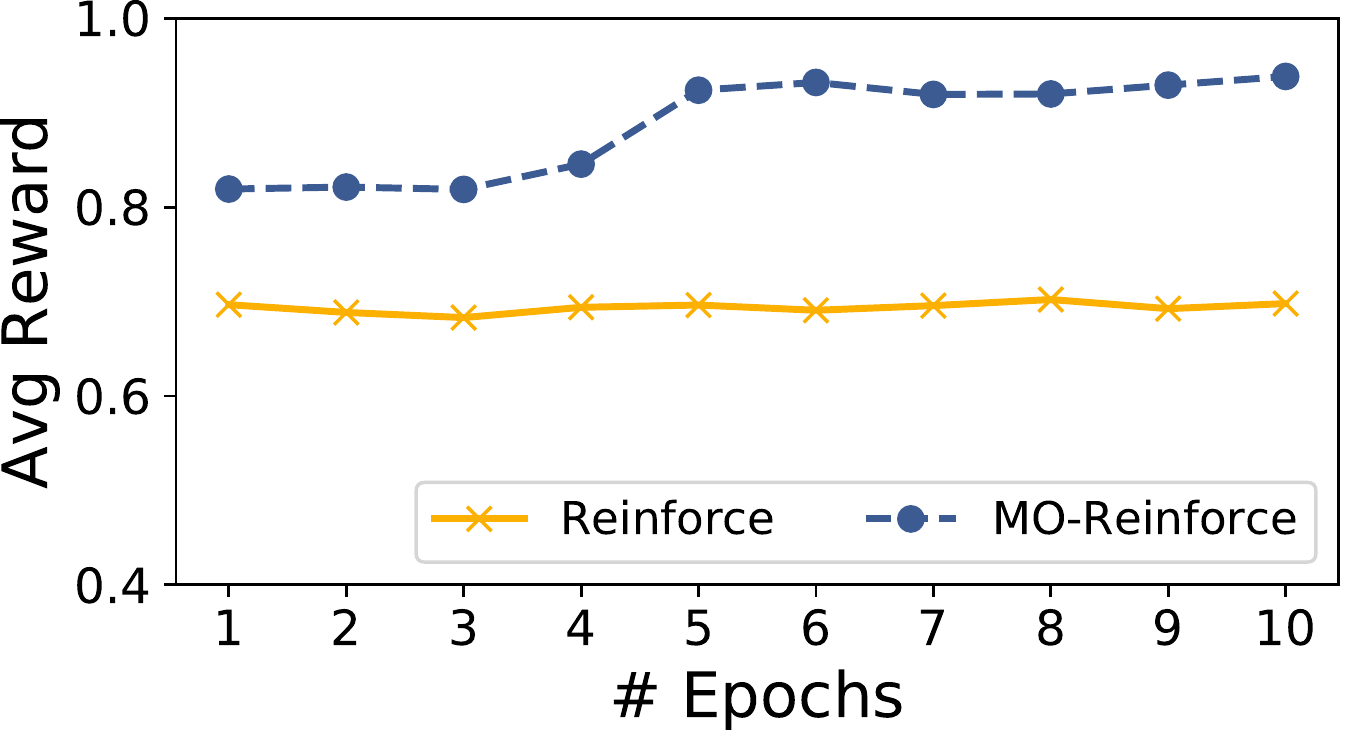}
\caption{Average rewards for \textit{Reinforce} and \textit{MO-Reinforce} candidates at each training epoch (It-En).}
\label{fig:avg_reward-ITEN}
\end{minipage}
\end{figure*}

\begin{figure*}
\centering
\begin{minipage}[b]{.46\textwidth}
\centering
\includegraphics[scale=0.5]{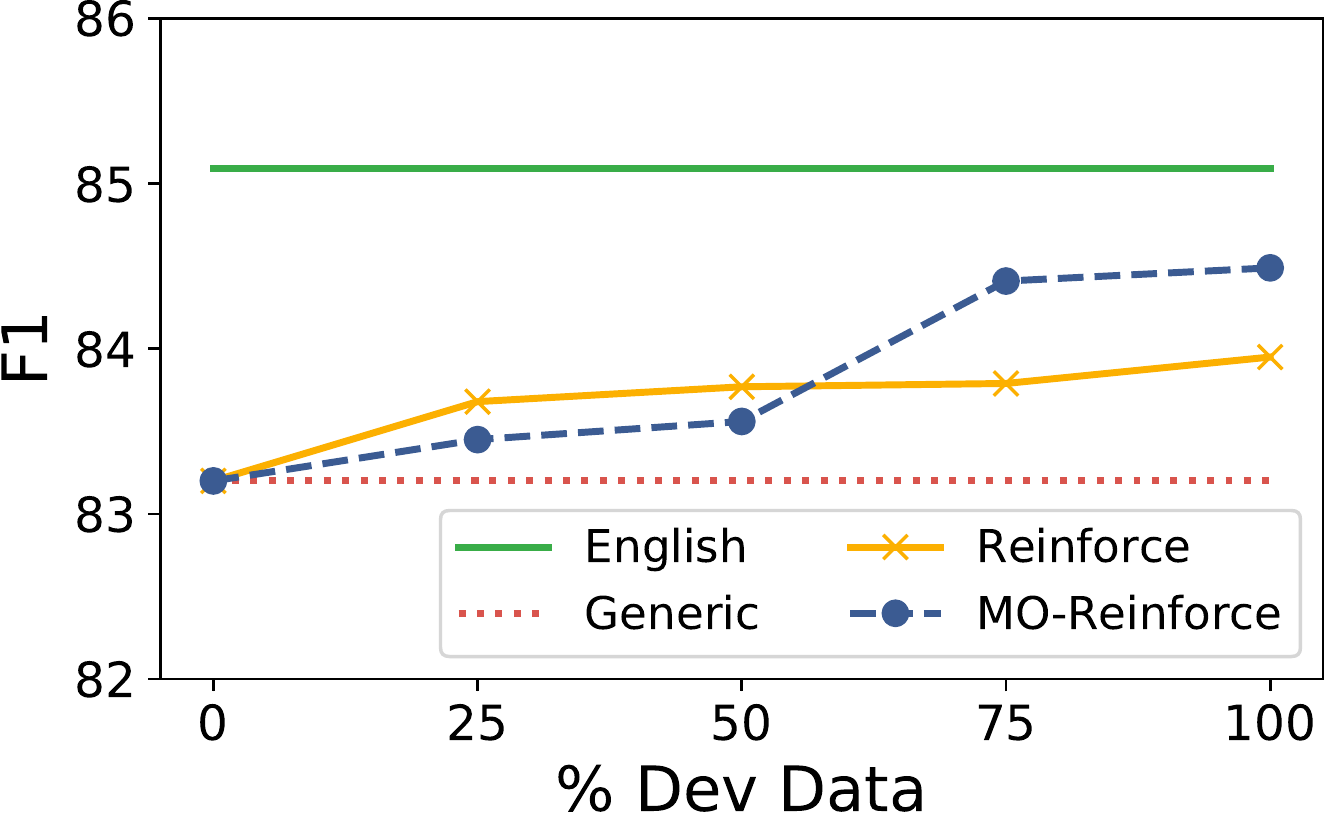}
\caption{Classification results with increasing amounts of labeled \mn{development} data (De-En).}
\label{fig:classif_results-DEEN}
\end{minipage}\qquad
\begin{minipage}[b]{.45\textwidth}
\centering
\includegraphics[scale=0.5]{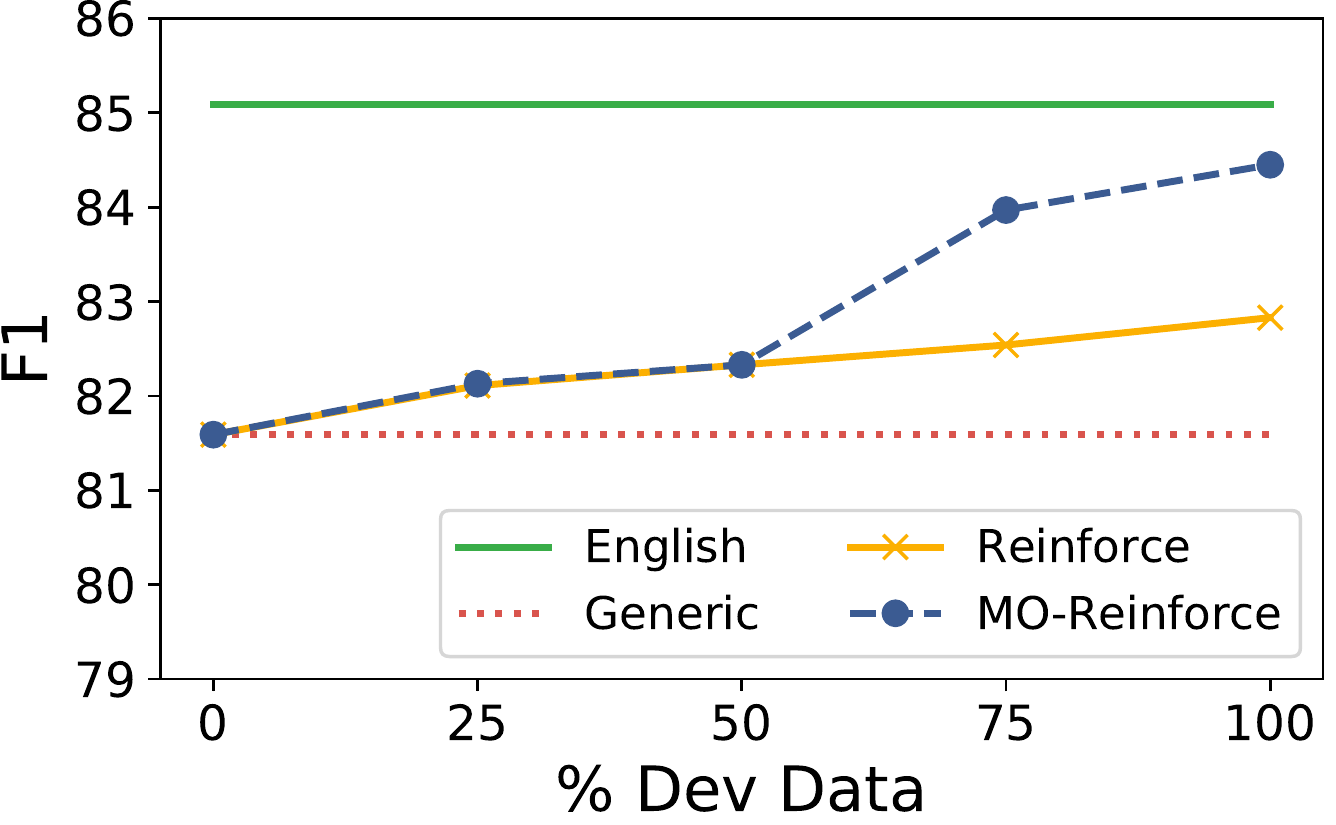}
\caption{Classification results with increasing amounts of labeled \mn{development} data (It-En).}
\label{fig:classif_results-ITEN}
\end{minipage}
\end{figure*}

Table \ref{tab:table1} shows our classification results,
presenting the F1 scores obtained by the different  MT-based approaches in the two  training conditions. 
When NMT is trained on 100\% of the parallel data, for both languages
\textit{Reinforce} produces translations that lead to classification improvements over those produced by the \textit{Generic} model (+0.5 De-En, +0.8 It-En). Although the scores are considerably better than those obtained by the \textit{Original} 
classifiers 
(+9.3 De-En, +7.2 It-En), the gap with respect to
the \textit{English} classifier
is still quite large (-1.4 De-En and -2.3 It-En).

The observed F1 gains over the \textit{Generic} model reflect an improvement in translation quality. Indeed, the BLEU score (not reported in the table) increases for both languages (+0.83 De-En, +1.37 It-En). 
As suggested by \newcite{kreutzer:2018}, this can be motivated by the fact that, rather than actually leveraging the feedback,  RL 
mainly benefits 
from the optimization 
on in-domain source sentences 
in the 
\mn{development} set.
This is confirmed by the fact that 
\textit{MO-Reinforce}, which uses a different sampling strategy, is able to outperform \textit{Reinforce} on both languages (+0.7 F1 on De-En, +1.7  on It-En), approaching the \textit{English} upper bound.

Unsurprisingly, 
the BLEU score obtained by 
\textit{MO-Reinforce}
is close to zero.
Indeed, the generated sentences are highly polarized and fluent, but not adequate with respect to the source sentence.  
For example, the positive instances in the test set are translated into  ``\textit{It's good!}'', ``\textit{I'm happy}'' or ``\textit{I'm grateful}'', and  the negative ones into ``\textit{it is not good!}'' or ``\textit{I'm sorry}''. 
This polarization shows that \textit{MO-Reinforce} 
maximizes the exploitation of the received sentiment feedback 
thus producing 
an output that, at the expense of adequacy, can be  easily classified by the downstream task.

When reducing the MT training data to only 5\%, emulating a condition of parallel data scarcity, all the classification results decrease (on average by 3-4 points). 
However, 
also in this case the output of \textit{MO-Reinforce} yields 
the closest scores to the \textit{English} upper bound.
This indicates that
our method does not require large data quantities to outperform the other approaches.

\begin{table}[t]
\centering
\begin{tabular}{l|l|l||l|l}
\toprule
& \multicolumn{2}{c||}{\textbf{De - En}} &\multicolumn{2}{c}{\textbf{It - En}}\\
      & \textit{5}\% &  \textit{100}\%  & \textit{5}\%  & \textit{100}\% \\ \midrule
\textbf{Generic}                  &  79.7  & 83.2  & 78.2    &81.6    \\ \midrule
 \textbf{Reinforce}                  &  80.4 & 83.7    & 77.8    & 82.8  \\
 \midrule
\textbf{MO-Reinforce}             &  \textbf{80.9} & \textbf{84.4}   & \textbf{80.3}    &\textbf{84.5}   \\  \midrule\midrule
 \textbf{English}               & 
 \multicolumn{4}{c}{85.1} \\
 \midrule
 \textbf{Original}                 & \multicolumn{2}{c||}{74.4} & \multicolumn{2}{c}{75.6}   \\
\bottomrule
\end{tabular}
\caption{Classification results (F1) obtained with: \textit{i)} automatic English translations by three  models 
(\textit{Generic}, \textit{Reinforce}, \textit{MO-Reinforce}),
and \textit{ii)} gold-standard  English  (\textit{English}) and \mn{untranslated} German/Italian  (\textit{Original}) tweets. 
}
\label{tab:table1}
\end{table}

\mn{To validate the hypothesis that \textit{MO-Reinforce} can better leverage the feedback from the downstream task,  Figures \ref{fig:avg_reward-DEEN} and \ref{fig:avg_reward-ITEN} show the average rewards for the De-En 
and the It-En 
candidates generated by \textit{Reinforce} and \textit{MO-Reinforce} at each epoch of the adaptation process. In line with the findings of \newcite{kreutzer:2018}, for both languages, \textit{Reinforce} is not able to leverage the feedback and to generate candidates that increase their reward during training.  
Indeed, its curves in the two figures show either a stable (It-En) or even a slightly downward trend (De-En) 
that confirms the known limitations of NMT to preserve sentiment traits of the source sentences (see Section \ref{sec:intro}). In contrast, the \textit{MO-Reinforce} curves show a clear  upward trend indicating a higher capability to exploit the feedback and produce, epoch by epoch, increasingly polarized translations that are easier to classify.}

\mn{Another important aspect to investigate  is the relation between the size of the \mn{development} set (i.e. the amount of human-annotated source language tweets 
needed by the RL methods) and final classification performance. Figures \ref{fig:classif_results-DEEN} and \ref{fig:classif_results-ITEN} 
show the De-En and It-En performance variations of \textit{Reinforce} and  \textit{MO-Reinforce} at different sizes of the development set. For comparison purposes, also the \textit{Generic} and \textit{English} results (which are independent from the 
\mn{development} set size) are  included. In both plots, each point is obtained by averaging the results of three different data shuffles. With limited amounts of data (25\%  and 50\%) \textit{Reinforce} and \textit{MO-Reinforce} have a similar trend. When adding more data, \textit{MO-Reinforce} shows a better use of the  labeled 
\mn{development} data, with a boost in performance that allows it to approach the \textit{English} upper bound in both language settings. These results suggest that the effort to create labeled data can be 
minimal, and with 75\% of the set (579 points) it is  \mn{already} possible to achieve a better performance than \textit{Reinforce}.}

\section{Conclusions}

\mn{We proposed  a novel interpretation of the  machine translation task, which pursues \textit{machine-oriented} quality criteria (generating translations that are best suited as input to a downstream NLP component) rather than the traditional \textit{human-oriented} ones (maximizing adequacy and fluency). We addressed the problem by adapting reinforcement learning techniques with a new, exploration-oriented sampling strategy that exploits  the results obtained on the downstream task as weak feedback.}
\mn{Instead of generating only one candidate totally randomly via multinomial sampling (i.e. random stepwise selection of each word during generation, as in \textit{Reinforce}), 
our approach selects $K$ full translation candidates, it computes the reward for each of them and finally chooses the one with the highest reward from the downstream task.
As shown by our experiments in sentiment classification,
this more focused (and application-oriented) selection allows our  ``MT for machines'' approach to: \textit{i)} better explore the hypotheses’ space, \textit{ii)} make better use of the collected rewards and eventually \textit{iii)} obtain better downstream classification results
compared to translation-based solutions exploiting either general-purpose models or previous reinforcement learning strategies.}
\mn{In future work, we will target new application scenarios, covering 
multi-class classification and regression tasks.}

\bibliography{emnlp-ijcnlp-2019}
\bibliographystyle{acl_natbib}

\end{document}